\documentclass[letterpaper, 10 pt, conference]{ieeeconf}  

\IEEEoverridecommandlockouts                              

\usepackage{graphics} 
\usepackage{epsfig} 
\usepackage{mathptmx} 
\usepackage{times} 
\usepackage{amsmath} 
\usepackage{amssymb}  
\usepackage{comment}
\usepackage{booktabs} 
\usepackage[table]{xcolor}
\usepackage{tikz}
\usepackage{pgfplots}
\usepackage{subcaption} 

\title{\LARGE \bf
Leveraging Event Streams with Deep Reinforcement Learning for End-to-End UAV Tracking
}
\author{Ala Souissi$^{1, 2}$, Hajer Fradi$^{1}$, Panagiotis Papadakis$^{1}$
\thanks{$^{1}$ IMT Atlantique, Lab-STICC, RAMBO team, France 
      {\tt\small {name.lastname}@imt-atlantique.fr}}%
\thanks{$^{2}$ Polytechnic School of Tunisia (EPT)
{\tt\small ala.souissi@ept.ucar.tn}}%
}

\begin{document}

\maketitle
\thispagestyle{empty}
\pagestyle{empty}

\begin{abstract}
In this paper, we present our proposed approach for active tracking 
to increase the autonomy of 
Unmanned Aerial Vehicles (UAVs)
using event cameras,  low-energy  imaging sensors 
that offer significant advantages in speed and dynamic range.
The proposed tracking controller is designed to respond to visual feedback from the mounted event sensor, adjusting the drone movements to follow the target.
To leverage the full motion capabilities of a quadrotor  and the unique properties of event sensors, we propose an  end-to-end deep-reinforcement learning (DRL) framework that maps raw sensor data from event streams directly to control actions for the UAV. 
 To learn an optimal policy under highly variable and challenging conditions, we opt for a simulation environment with domain randomization for effective transfer to real-world environments.  
We demonstrate the effectiveness of our approach through experiments in challenging scenarios, including fast-moving targets and changing lighting conditions, which result in improved generalization capabilities.

\end{abstract}

\section{INTRODUCTION}
 The technology of UAVs, known as drones, has been  increasingly used in humanitarian missions for search and rescue, surveillance for safety control and emergency contingency plan, or for  guiding tasks \cite{rejeb2021humanitarian,fang2024strategies,9543551,9321707}.
 This rising interest emphasizes the need for autonomous navigation rather than manual control.
Active tracking is one of the most crucial tasks for UAVs, requiring the tracker to keep the target centered in its field-of-view (FOV), 
 relying mostly on visual observations to follow the moving target \cite{ma2023target}. These tracking capabilities are useful in many applications including search and rescue missions, where a piloted drone can lead exploration while a fleet of autonomous drones navigates autonomously in GPS-denied environments by following the leader \cite{heintzman2021anticipatory, xu2023collaborative}.


Active tracking with drones presents significant challenges due to the complexity of the tracking process and the nonlinearity of the system. Drone dynamics are affected by uncertain environmental conditions and nonlinear effects from aerodynamic forces, torques, payload variations, and control signal noise. In dynamic environments, where target motion and conditions change rapidly, maintaining accurate tracking becomes even more difficult. 
These challenges make it difficult to design a system controller that can reliably maintain stable tracking.

Earlier works have mostly employed Proportional-Integral-Derivative (PID) control \cite{Wang_2022} or Linear Quadratic Regulator (LQR) \cite{MARTINS2019176} approaches.
However, designing such controllers requires defining a highly accurate model of drone dynamics, which is often difficult to obtain.  
Recent advances in the field involve using reinforcement learning; however, like classic controllers,  
it mostly requires access to the  target state \cite{xi2021anti}.
Accurately estimating that while both the tracker and the target are in continuous motion is challenging and requires integrating a high-performance module for precise target localization.  

To address the aforementioned challenges, we opt for designing an end-to-end deep reinforcement learning-based controller that can process raw sensor data 
instead of training detection-based policy. 
A DRL-based tracker adapts to environmental changes by learning from past experiences, enabling model refinement without altering system components, unlike classic controllers.
Additionally, a highly accurate dynamic model is not required because the DRL algorithm can learn through data, even when the model does not perfectly capture all the complexities of the control system.

\begin{figure}[!t]
    \centering
    \includegraphics[width=0.4\textwidth]{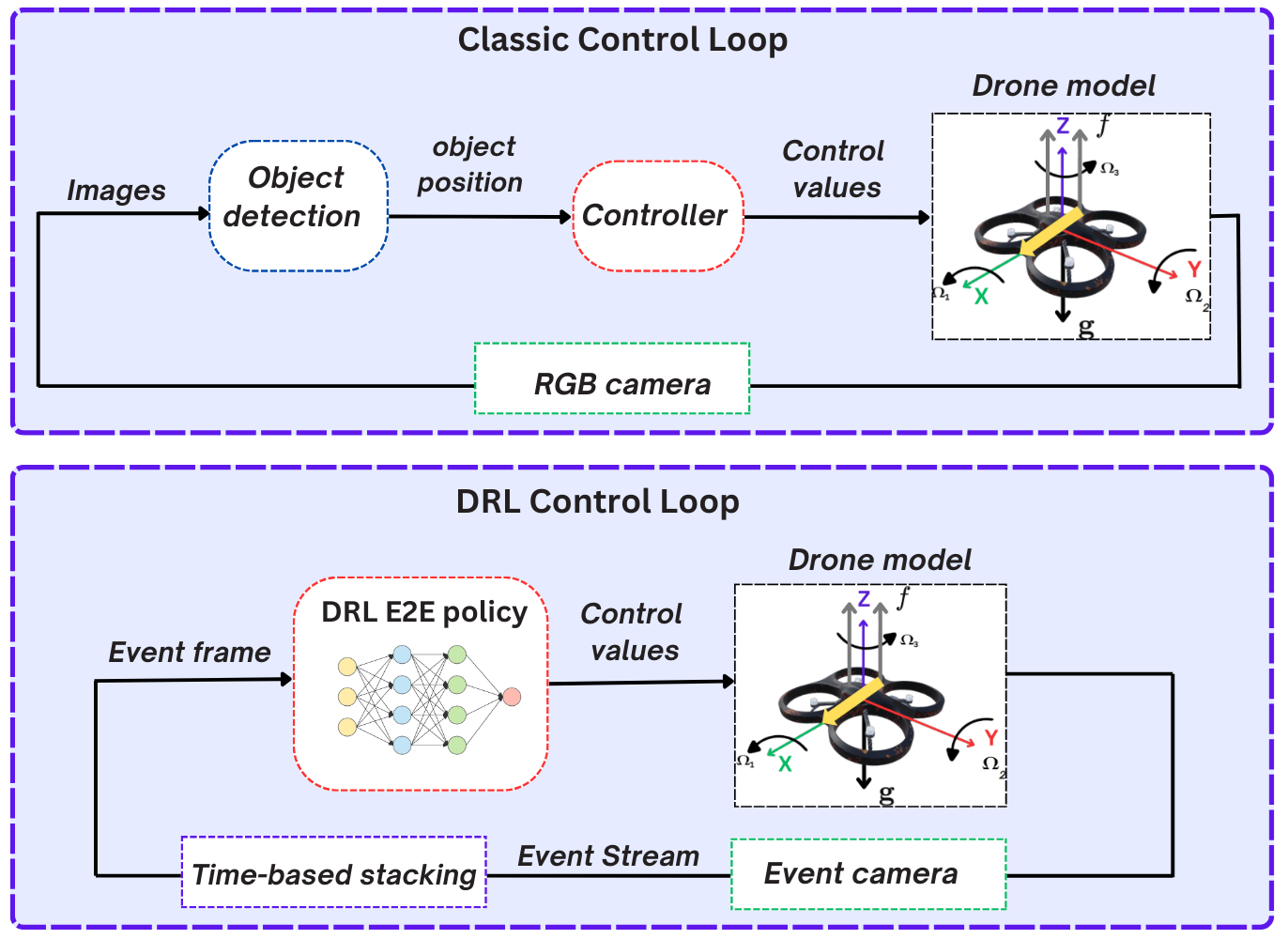}
    \caption{DRL controller  using event streams vs. classic controller for UAV  active tracking.}
    \label{fig:motiv}
\end{figure}

Unlike previous studies that  have shown  the potential of reinforcement learning in control tasks \cite{zhao2022deep,trehan2022towards,devo2021enhancing}, the proposed end-to-end framework not only bypasses the detection task but also  leverages the unique properties of event sensors over conventional RGB cameras. The latter, with their low frame rates and limited dynamic range, pose significant challenges in control.
Motivated by the recent success of event cameras in vision applications \cite{gallego2020event},  we aim  to  instead investigate these newly emerging bio-inspired sensors. 
Thanks to their distinctive properties, such as high dynamic range, high temporal resolution, and low latency, a large interest has been shown  in exploiting these new sensors, 
especially in autonomous vehicles where rapid responses, adaptability to weather and lighting changes, and robust visual information at high speeds are crucial \cite{gehrig2021dsec}.
An illustration of the comparison between a classic controller and our proposed end-to-end DRL-based controller processing events is shown in Fig. \ref{fig:motiv}. 

Training a DRL model involves trial and error where the drone learns an optimal strategy based on previous trials. Since this requires massive training, which is not feasible with a real drone, due to the risk of drone crashes,  we opt for simulated aerial data collection which allows numerous simulated trajectories.
This simulated environment serves as a crucial resource for training our proposed DRL-based controller and is complemented by domain randomization for effective real-world transfer and parallel training to manage complexity.
\\
\\
Our contributions in this paper are summarized as follows:
\begin{itemize}
\item 
To enable autonomous active tracking, we propose a  DRL-based UAV controller designed for aerial-to-aerial tracking. This controller maps visual information from mounted sensors to control drone  commands using an end-to-end architecture.
\item 
For accurate tracking under challenging situations, we integrate event cameras with the DRL framework to overcome RGB camera limitations, applying this combination to aerial-to-aerial tracking for the first time,  to the best of our knowledge.
\item 
To minimize drone damage and maximize exploration, the training is conducted in a simulated environment. For effective real-world transfer, the training is complemented  by domain randomization and parallel training to manage complexity.
\item Using  environments from the DARPA Subterranean Challenge \cite{darpa2024subtchallenge}, the proposed end-to-end architecture achieves improved performance in scenarios involving high-speed targets and changing lighting conditions.
\end{itemize}

\section{Related work}
\label{sec:rel}
\subsection{Visual Tracking}
In visual tracking, a key distinction is made between passive and active tracking.
In passive tracking, the object position is estimated in each frame based on its previous state. 
Object detection algorithms, such as YOLO \cite{wang2024yolov9}, SSD \cite{Liu_2016}, or Faster R-CNN \cite{ren2016fasterrcnnrealtimeobject}, are used to locate the object. Once identified, tracking involves maintaining the object position in subsequent frames using methods like the Kalman filter, Mean-Shift, CAMShift \cite{vaitheeswaran2015leaderfollowerformationcontrol}, or deep learning-based trackers like Deep-SORT \cite{wojke2017simpleonlinerealtimetracking}. The camera is typically fixed, so the tracking focuses on processing stationary frames.
Unlike passive tracking, active tracking does not require explicit object localization \cite{arakawa2020exploration,10493111,luo2018endtoendactiveobjecttracking}. 
Instead, feedback from previous camera frames is used to adjust camera orientation and position enabling to focus on the target in dynamic and responsive way to the target movement.
In the case of UAV tracking, the drone receives visual information from mounted cameras and adjusts its position and orientation to keep the target centered in its FOV. 
\subsection{Active Drone Tracking}
For  active tracking, some related works rely on classical controllers \cite{9097927,Wang_2022, trehan2022towards}. 
For instance, in \cite{trehan2022towards}, 
predictive learning is  combined with reactive control systems to perform self-supervised active action localization. 
The reactive control, inspired by PID controllers, adjusts camera orientation to keep the target within the FOV.
Recent advances in the field have shown the potential of using DRL in control tasks, but most approaches assume access to the target position and are designed for either ground-to-ground or air-to-ground tracking \cite{devo2021enhancing,zhao2022deep}.
In \cite{devo2021enhancing},
a DRL-based visual active tracking system that provides continuous action policies is proposed, however, experiments are conducted on ground mobile robots which are controlled via discrete actions. 
The method in \cite{zhao2022deep} performs aerial-to-ground tracking using a policy learned from training to fly toward a fixed target.

We note that most DRL-based controllers have been implemented for ground robots or for tracking moving objects on the ground, while aerial-to-aerial tracking remains less explored. Furthermore, the proposed approaches are constrained by the limitations of conventional RGB cameras, which, with their low frame rates and limited dynamic range, present significant challenges for dynamic and accurate control. While event cameras have been investigated for UAV applications, primarily for obstacle avoidance \cite{vemprala2021representation, arakawa2020exploration}, to the best of our knowledge, our work is the first to explore their use for UAV active tracking.

\section{Preliminary definitions}
\subsection{Dynamic Quadrocopter Model}
The quadrotor model, inspired by the study of Mark \textit{et al.} in \cite{7299672}, represents the drone as a rigid body with six degrees of freedom, namely, three translational and three rotational along the 3D body axes. 
This model is controlled by a scalar thrust \( f \) representing the total thrust value along the Z axis, 
and angular velocity expressed in the fixed body frame as \( \boldsymbol{\Omega} = (\Omega_1, \Omega_2, \Omega_3) \). 

The quadrotor state consists of the position, velocity, and orientation. The differential equations are as follows:
\begin{equation}
\ddot{\mathbf{p}}(t) = \mathbf{R}_3(t) \frac{f(t)}{m} + \mathbf{g},
\label{eq:diff_equation_1}
\end{equation}
\begin{equation}
\dot{\mathbf{R}_3}(t) = \mathbf{R}(t) [\boldsymbol{\Omega}(t)]_{X},
\label{eq:diff_equation_2}
\end{equation}
\begin{equation}
\boldsymbol{\Omega}_\times = \begin{bmatrix} 
0 & -\Omega_3 & \Omega_2 \\ 
\Omega_3 & 0 & -\Omega_1 \\ 
-\Omega_2 & \Omega_1 & 0 
\end{bmatrix},
\end{equation}
where $\mathbf{p}(t)$, $\mathbf{R}(t)$, and $\boldsymbol{\Omega}(t)$ are the absolute position, orientation, and angular velocity of the tracker drone at time $t$, respectively. The term $\mathbf{R}_j(t)$ refers to the $j$-th column of the orientation matrix $\mathbf{R}(t)$. The total thrust is denoted by $f(t)$, with $m$ representing the drone mass. The gravity vector is given by $\mathbf{g} = [0 \; 0 \; -9.8]^\top \, \text{m/s}^2$, and $[\boldsymbol{\Omega}(t)]_{X}$ denotes the skew-symmetric matrix associated with $\boldsymbol{\Omega}(t)$.
 The thrust value \(f(t) \) and the angular rates are constrained within the following ranges:
\begin{equation}
0 \leq f_{\text{min}} \leq f \leq f_{\text{max}},
\label{eq:thrust_constraints}
\end{equation}
\begin{equation}
-\Omega_{\text{max}} < \Omega_i < \Omega_{\text{max}} \quad \text{for} \quad i \in \{1, 2, 3\},
\label{eq:angular_rate_constraints}
\end{equation}
where the limit values \( f_{\text{min}} \), \( f_{\text{max}} \), and \( \Omega_{\text{max}} \) can be set to match certain drone specifications. 
\subsection{Reinforcement Learning-based Tracking}
For UAVs, the active tracking problem involves using visual sensory data to generate actions that keep the target centered and maintain a certain distance.
We formulate the UAV active tracking by reinforcement learning as a Partially Observable Markov Decision Process (POMDP) \cite{LITTMAN1994157}, an extension of Markov Decision Process (MDP). POMDP is defined by the tuple \((\mathbb{S}, \mathbb{A}, \mathbb{O}, T,  R, Z, \gamma)\), where \(\mathbb{O}\), \(\mathbb{S}\), and \(\mathbb{A}\) are the observation
, the state, and the action spaces
, respectively. The discount factor \(\gamma\) balances immediate and future rewards. \(T\) is the transition probability to a new state. \(Z\) defines the probability of obtaining the current observation given the current state and  \(R\) is the reward function.

At each time step, the agent (UAV) interacts with its environment as follows: (i) the agent receives a visual observation \(o_t \in \mathbb{O}\); (ii) based on this observation and using a stochastic policy \( a_{t} \sim \pi(a_{t} \mid o_{t}) \), the agent selects actions; (iii) the agent gets an immediate reward \(r_{t+1} = R(s_{t+1})\) as a function of the new state and receives a visual observation \(o_{t+1}\) correlated with the new state \(o_{t+1} \sim {Z}(o_{t+1} \mid s_{t+1})\). 
This formulation is used to find the optimal policy  \(\pi^*\) that maximizes the expected cumulative reward over interactions with the environment. In the following, we instantiate the observation, state, and action spaces for the problem under consideration.
\\
\textbf{Observation Space: }
The observation  derived from visual sensors is defined as a sequence of the \( N \) latest images: \( O(t) = (I(t-N+1), ..., I(t)) \), where \( I(t) \) is the current image. The observation space is therefore defined as:
$\mathbb{O} = (H, W, C)^N$,  with $W\times H$ is the spatial resolution and $C$ is the number of channels.
\\
\textbf{Action Space:} 
The actions are the control decisions for tracking, defined  as \(a_t = (f(t), \Omega(t))\). 
The action space is continuous and is defined as:
$\mathbb{A} = \underbrace{[0, f_{\text{max}}]}_{\text{thrust}} \times \underbrace{[-\Omega_{\text{max}}, \Omega_{\text{max}}]^3}_{\text{angular rates}} \subset \mathbb{R}^4$.
\\
\textbf{State Space: }
The state space of the system at time \( t \) is defined as \( s_t = (P_t, V_t, A_t) \in \mathbb{R}^9 \), with \( P_t = (x(t), y(t), z(t)) \) is the relative position of the target along three axes with respect to the tracker, \( V_t = (v_x(t), v_y(t), v_z(t)) \) is the relative velocity, and \( A_t = (a_x(t), a_y(t), a_z(t)) \) is the relative acceleration. 

\section{Method}
\label{sec:prop}

\subsection{Asymmetric Soft Actor-Critic Framework}
\begin{figure}[htb]
    \centering
    \includegraphics[width=.37\textwidth]{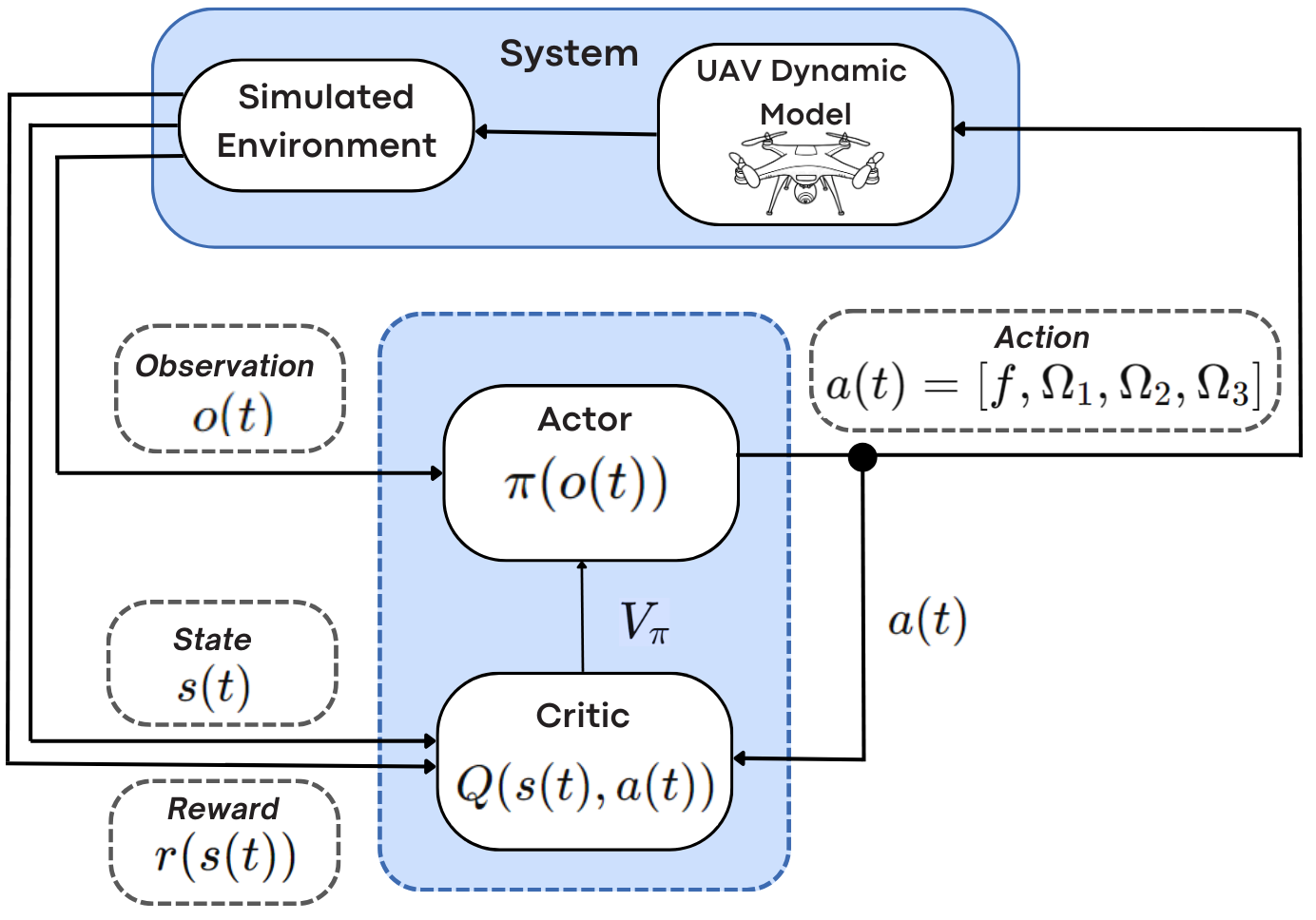}
    \caption{The flowchart of Asymmetric Soft Actor-Critic (ASAC) learning framework: the actor and critic networks work together to optimize the policy.}
    \label{fig:asac_learn_framework}
\end{figure}

The learning framework represents an asymmetric modification  of the Soft Actor-Critic  algorithm as shown in Fig. \ref{fig:asac_learn_framework}.

\subsubsection{Asymmetric Actor-Critic} 
The actor uses a stochastic policy to generate actions from the current observation, while the critic evaluates these actions with a Q-function based on the selected action and the current state. Both are modeled with neural networks and trained together to optimize the policy.
In soft actor-critic, the optimal policy maximizes both expected rewards and entropy, encouraging the agent to explore new strategies while balancing exploration and exploitation.

The characteristic of asymmetry leads in using different inputs for the actor and critic. In robotics, the policy often relies on partial observations from sensors, which provide a noisy and incomplete view of the environment.
However, the critic 
has access to the full system state in the controlled training environment, optimizing the learning process despite the robot restricted sensory input.
This modified Actor-Critic algorithm with asymmetric inputs shows promise, as the critic has access to task-relevant information and a full system view during training speeds up convergence, improving overall learning efficiency.
    \begin{figure*}[htb]
    \centering
    \includegraphics[width=0.9\textwidth]{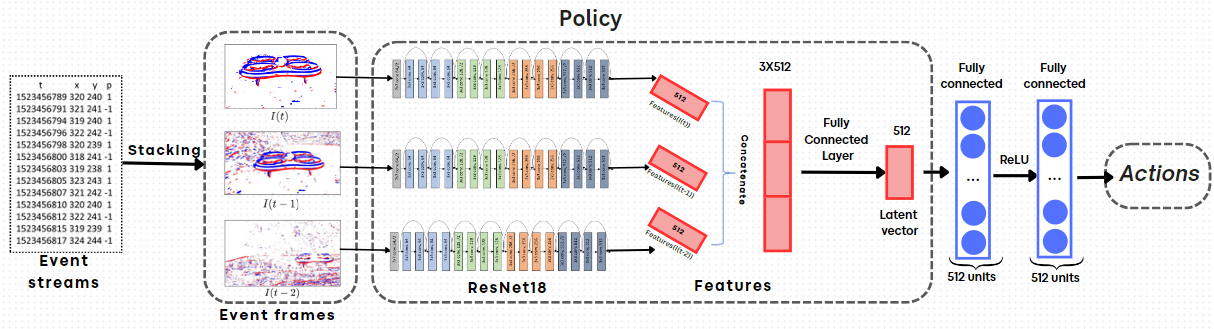}
    \caption{Overview of the proposed end-to-end architecture processing events for UAV active tracking.}
    \label{fig:e2e}
\end{figure*}
\subsubsection{Reward shaping}
The main control objective is to maintain the target in a defined relative position  $P_t = (x(t), y(t), z(t))$ with respect to the tracker. 
Specifically, we aim that the target position along the \( x \)-axis of the tracker body frame to be equal to an optimal distance \( d^* \), and  along  the \( y \) and \( z \) axes equal to 0  in order to keep the target centered in the FOV of the tracker, as shown in Figure \ref{fig:motiv}, which illustrates the drone's body frame and axis.

To achieve this, we shape the reward function based on the relative position $P_t$.  
The reward function \( r_e(t) \) is defined as the cubic root of the product of three reward components at each axis:
\begin{equation}
r_e(t) = \sqrt[3]{r_x(t) \cdot r_y(t) \cdot r_z(t)}
\end{equation}
where:
$$r_x = \max \left(0, 1 - \left| x(t) - d^* \right|\right)$$ $$r_y = \max \left(0, 1 - \frac{2}{\pi} \cdot \arctan \left(\frac{y(t)}{x(t)}\right) \right)$$ 
$$r_z = \max \left(0, 1 - \frac{2}{\pi} \cdot \arctan \left(\frac{z(t)}{x(t)}\right) \right)$$
The product of the three reward functions promotes  maintaining visual contact at  the optimal distance. If any of these conditions is not met, the total reward is  set to zero.

An additional penalty term is included to optimize the UAV  linear velocity, discouraging excessive speeds that could lead to instability or inefficient tracking behavior, defined as follows:
\begin{equation}
r_v(t) = \frac{v(t)}{1 + v(t)}
\label{eq:rv}
\end{equation}
To avoid collisions, we introduce a penalty term: when the relative distance falls below the limit  \( \|P(t)\| < d_{min} \), it incurs a large negative value.
These  reward terms are combined in an  overall reward function $r(t)$ defined as: 

\begin{equation}
r(t) = \begin{cases}
r_e(t) - \alpha r_v(t), & \text{if } \|P(t)\| > d_{min} \\
-k_c & \text{otherwise}
\end{cases}
\label{eq:r}
\end{equation}
where \( k_c \) is a large positive constant, and \( \alpha \) is a positive weighting parameter. 

\subsection{End-to-End  Event-based Tracking Architecture}
 Within the presented ASAC framework, we train a single end-to-end model to generate actions directly from the raw event data.
 The overall architecture is shown in Fig. \ref{fig:e2e}.
\subsubsection{Event Data Processing}
Compared to conventional cameras capturing images at a fixed frame rate, event cameras respond to brightness changes for every pixel asynchronously and independently.
This results in a stream of events that are spatially sparse and asynchronous. Each event is a tuple  \( e = (t, x, y, p) \), where \( t \) is the timestamp  at which the event is triggered, $(x,y)$ are the spatial coordinates, and $p$ is the polarity indicating the sign of the change. To align with conventional image-based vision, event streams need to be transformed into a 2D spatial grid representation. A common way to represent events is the stacking on time which involves incorporating a sequence of events \( E = \{e_i \mid t \le i < t + \Delta t\} \)  within a time
interval  \( \Delta t \), resulting in an event frame.
To train our policy, the observation is defined as a sequence of the \( N \) latest event frames ($N=3$ in our case): \( O(t) = (I(t-2), I(t-1), I(t)) \). 
To process event frames and capture task-relevant features, the policy architecture is designed as follows. 

\subsubsection{Neural Networks of Actor and Critic }
Actor and Critic are modeled with the following networks.  
\\
\textbf{ Actor-NN:} 
The deep neural network defining the actor consists of two blocks. The first block is a  feature extractor that maps observations   to a feature vector, capturing the spatial and temporal information. The feature embedding space has a higher correlation with the state space, giving a much more informative representation of the current state.
The event frames, which serve as  observations, are processed using ResNet18 \cite{he2015deepresiduallearningimage} as  feature extractor due to its balance of complexity and performance. 
The feature extractor has as input a sequence of  3 event frames and maps each frame to a single feature vector of dimension 512. 
The three vectors are then concatenated along the first axis to obtain a single latent vector. We then use a fully connected layer to reduce the vector dimension from $3 \times 512$ to 512.
 The second block consists of two linear layers with 512 neurons each followed by a \texttt{tanh} activation. It processes the features of dimension 512 and produces the mean and log standard deviation for the Gaussian probability distribution over the possible actions.
\\
\textbf{Critic-NN:}  The critic network receives the full state of the environment, represented by a nine-dimensional vector \( s_t = (P_t, V_t, A_t) \in \mathbb{R}^9 \) and the chosen action \( a(t) \). The critic is designed using a straightforward architecture based on fully connected neural network layers. First, we use a flatten layer to transform the input into a one-dimensional vector. The main part of the architecture consists of three dense layers. The first layer maps the input to 512 neurons, followed by a second hidden layer with 512 neurons. The final output layer produces a single scalar value, representing the estimated action value \( Q^\pi(s(t), a(t)) \). A \texttt{tanh} activation function is applied at the output to constrain the value range.


\section{Experiments and Results}
\label{sec:exp}
\subsection{Experimental Setup}
Since training a policy via deep reinforcement learning would require a massive number of trial and error attempts, which is impractical with a real drone, we opt for a simulated environment that provides realistic and infinite training data.
In particular,  we build a simulated environment using AirSim simulator \cite{shah2018airsim} that supports aerial vehicles and Unreal Engine (UE)  \cite{sanders2016introduction} as graphics engine offering high-fidelity physics and realistic rendering with highly detailed visualizations. 
AirSim  provides flexibility in drone dynamics modeling by supporting both internal and external dynamic models. This enables to leverage the previously defined drone dynamic model while using the simulator  for accurate sensor feedback based on the drone position and orientation.

\subsection{Training and Evaluation Environments}
We generate a training environment called the \textit{box} environment, where the drone learns optimal tracking strategies under highly variable conditions. To favor smooth transfer to real-world environments, we employ domain randomization \cite{horvath2022object}, introducing enough variability. 
This approach helps the model handle changes and maintain policy effectiveness across different scenarios.
More precisely, the training environment includes random variations in wall textures, ranging from simple to highly patterned designs, as well as changes in color, light intensity, and orientation.

\begin{figure}[htb]
    \centering
    \begin{subfigure}[b]{0.49\textwidth}
        \centering
        \includegraphics[width=0.8\textwidth, height=1.7cm]{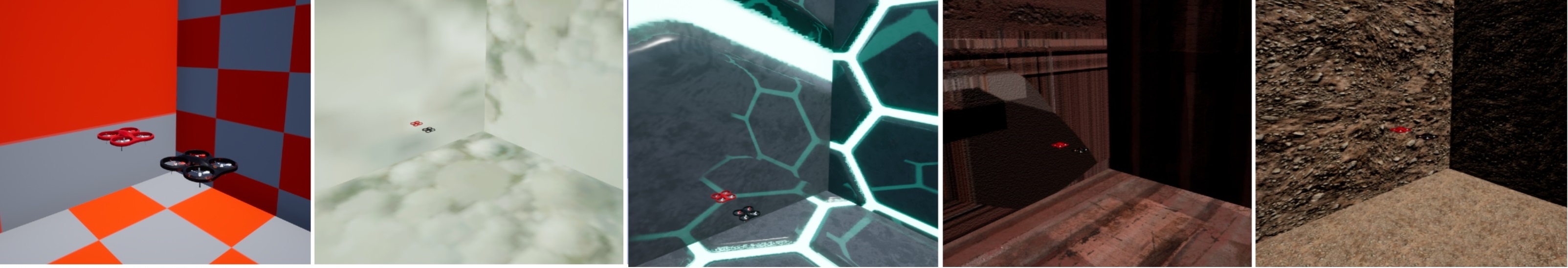}
        \label{fig:training_env}
    \end{subfigure}
    \begin{subfigure}[b]{0.49\textwidth}
        \centering
        \includegraphics[width=0.8\textwidth, height=1.7cm]{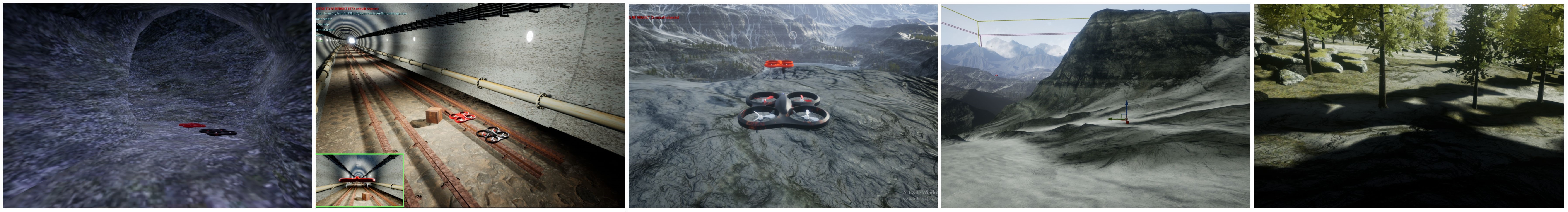}
        \label{fig:DAPRA_env}
    \end{subfigure}
    \caption{Examples of training \textit{box} environments with varying textures (first row) and DARPA environments used for evaluation (second row).}
    \label{fig:environments}
\end{figure}

 To evaluate the trained model, we use environments \cite{nguyen2020autonomousnavigationcomplexenvironments} inspired by the DARPA Subterranean Challenge\footnote{https://github.com/osrf/subt/tree/master/subt\_ign/worlds} \cite{darpa2024subtchallenge}, focusing on robot navigation in complex  backgrounds such as caves, urban scenes, tunnels, and mountains, all of which are entirely new to the policy.
 These environments are challenging due to low lighting and high-texture scenes, making them ideal for evaluating our policy. We further test the model in various \textit{box} environments to assess its generalization capabilities. 
For both cases, we test our policy  under changing lighting conditions and scenarios involving tracking very fast targets. 
Additionally, trajectories are randomly generated so as to ensure that the target moves in all directions and periodically pauses at random intervals.
Some examples of the \textit{box}  and DARPA environments are shown in Fig. \ref{fig:environments}.

\subsection{Comparisons and Implementation Parameters}
The training was conducted using an NVIDIA RTX A4500 GPU, an i7 12-core CPU, and 32 GB of RAM. To accelerate the training process, we utilize parallel training with 7 agents. 
The proposed end-to-end event-based UAV tracking is compared with two settings. The first is a baseline method that trains the policy based on object detection, where the observation space consists of target positions and distances. The policy architecture is defined as a simple MLP, with the flattened detection information as input, followed by three fully connected layers with ReLU activation. The second setting is an RGB-based approach, where the perception input is replaced by the latest RGB images instead of event data, while the other architectural details remain unchanged.

In  all settings, we defined a total number of epochs equal to 70 epochs, each one involving 50,000 time steps, with evaluation episodes set to 6. The buffer size is maintained at 10,000, and the batch size is set to 64. Training occurs every 8 timesteps, with a learning rate of 0.0003 and a discount factor of 0.99.
In the reinforcement learning environment,  each episode lasts 40 seconds, with the optimal tracking distance $d^*$ being 0.2 meters. The action space includes angular rates between \([-3.5, 3.5]\) rad/s and thrust between \([-18, 18]\) N. 
Reward parameters are the penalty weight \(\alpha\) equal to  0.4 and the reward penalty constant \(k_c\) equal to  10.
Events are stacked at a time interval $\Delta t$=0.005 seconds.


\subsection{Training Process}
To ensure that each training episode has a unique trajectory and expose the tracker to new scenarios, the target motion is generated using random movements and velocities, ranging from slow to fast, with sinusoidal trajectories. We use random amplitudes, phases, and frequencies sampled from a defined interval. 
We introduce short random periods where the target stops moving, allowing the model to encounter and adapt to static scenarios.

\begin{figure}[htb]
    \centering
    \includegraphics[width=0.3\textwidth]{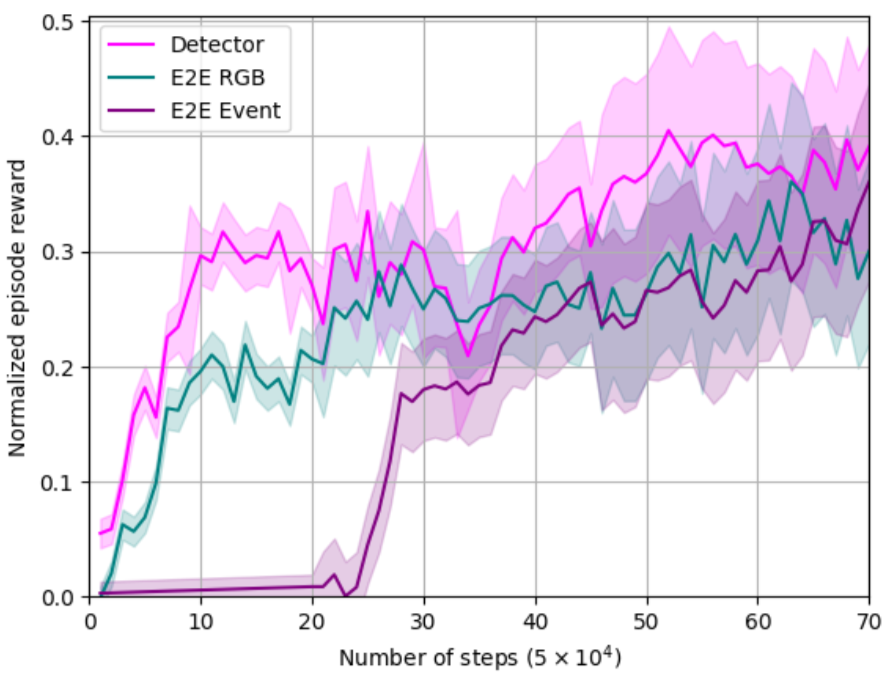}
    \caption{Learning plots  of the  detection-based (baseline), end-to-end RGB-based (E2E RGB), and end-to-end event-based (E2E event) approaches.}
    \label{fig:Training_performance}
\end{figure}

During training, the drone follows the target for up to 40 seconds per episode. 
The episode ends if the drone collides with the target, loses sight of it, or reaches the time limit. This strategy ensures that  the target stays within the field of view and at a convenient distance to optimize the policy.
After each training epoch, the policy is evaluated across 6 different episodes using the mean and the standard deviation of the cumulative reward per episode. 
This process is repeated 70 times until the final trained policy is obtained. 
 The mean cumulative reward per episode curves of the training phase are shown in Fig.\ref{fig:Training_performance}. 

For the object detection-based approach, the agent learns faster due to the low-dimensional vector input and the simplicity of the policy architecture. The learning curve stabilizes around 0.35, achieving a maximum mean reward value of \(0.4050 \pm 0.09\). It is important to note  that the object detector-based policy is based on the  ground truth detections. However, it shown at the end of the training plots, that the rewards of both end-to-end approaches (RGB and event), are very close to it with event-based method performing better.
\subsection{Limitations of Object Detection-based Policy}
\begin{figure}[htb]
    \centering
    \includegraphics[width=0.5\textwidth]{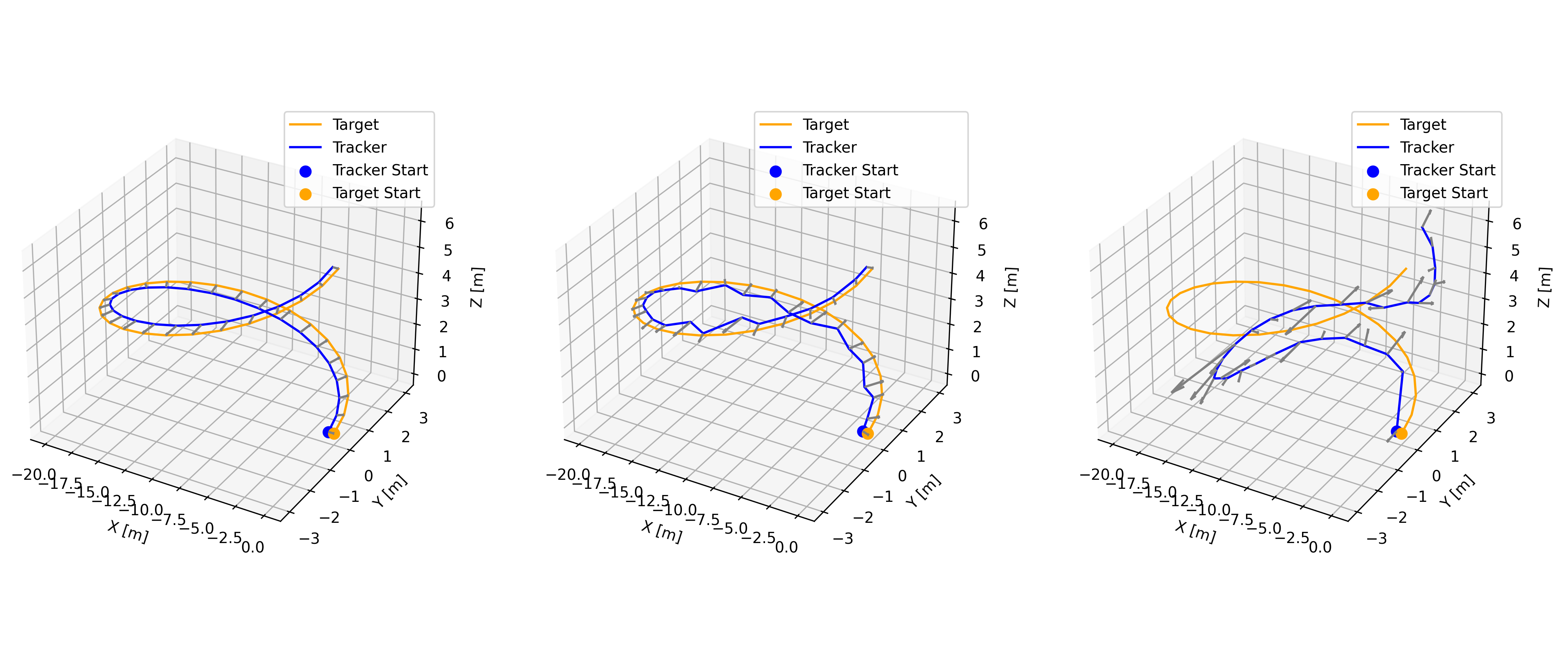}
    \caption{Example of the tracker and the target trajectories using object detector-based policy with different noise values: \(\eta = 0\), \(\eta = 0.06\), and \(\eta = 0.12\).}
    \label{fig:OD_adding_noise}
\end{figure}

To validate the effectiveness of the end-to-end approach, we evaluate the performance of object detector-based policy trained on ground truth detections. Using an object detector can simplify the learning process, allowing the agent to learn faster, as discussed in the previous section. However, a drop in detection accuracy  can significantly reduce the overall system performance. In Fig. \ref{fig:OD_adding_noise}, we illustrate the impact of adding noise to the detections by comparing the trajectories of the target and the tracker.

In this example, the target trajectory is fixed, and only the noise weight value is changed. 
As expected, the figure shows a drop in overall tracking  performance as the noise increases. 
At \(\eta = 0.06\), the tracker struggles slightly, indicating medium noise. At 
\(\eta = 0.12\), the tracker fails due to the high noise.
This demonstrates how even a slight change in object detector performance can significantly impact the overall tracking system, emphasizing the advantage of an end-to-end architecture that learns directly relevant features from raw inputs within the DRL framework.
During training, the reward system encourages the agent to encounter scenarios relevant to the tracking task, enabling the feature extractor to focus on task-specific data during optimization. Jointly optimizing the feature extractor and the network head can enhance overall performance. 

\subsection{Results and Analysis}
After discussing the limitations of the detection-based policy, we  compare, at this stage, our proposed approach to the RGB-based approach, where event inputs are substituted with RGB images.
For this comparison, we use ASAC as the baseline reinforcement learning algorithm with two types of perception inputs: RGB images and event frames. Both methods use the same convolutional neural network architecture, ResNet-18, with identical parameter initialization for fair comparison. Additionally, the task predictive head of the network remains unchanged. 
Both policies are trained under the same conditions in the \textit{box} environment.
Table \ref{tab:performance_comparison} presents the evaluation results of both approaches. 

\begin{table}[t]
\fontsize{8}{8}\selectfont
\renewcommand{\arraystretch}{1}
  \centering
  {\small{
  \begin{tabular}{@{}lcc@{}}
    \toprule
    \multicolumn{1}{c}{Scenarios} & E2E Event& E2E RGB \\
    \midrule
    \textit{Box} environment (Low-light) & \cellcolor{gray!20}0.45 \ensuremath{\pm} 0.20 & 0.35 \ensuremath{\pm} 0.22 \\
     \textit{Box} environment (Fast target) & \cellcolor{gray!20}0.47 \ensuremath{\pm} 0.19 & 0.34 \ensuremath{\pm} 0.21 \\
    \textit{Box} environment (Normal) & 0.50 \ensuremath{\pm} 0.10 & \cellcolor{gray!20}0.52 \ensuremath{\pm} 0.09 \\
    \hline
    DARPA environment (Low-Light) & \cellcolor{gray!20}0.28 \ensuremath{\pm} 0.17 & 0.19 \ensuremath{\pm} 0.16 \\
    DARPA environment (Fast target) & \cellcolor{gray!20}0.26 \ensuremath{\pm} 0.12 & 0.22 \ensuremath{\pm} 0.08 \\
     DARPA environment (Normal) & 0.28 \ensuremath{\pm} 0.11 & \cellcolor{gray!20}0.48 \ensuremath{\pm} 0.09 \\
    \bottomrule
  \end{tabular}
  }}
  \caption{Performance comparison of event and RGB policies across different scenarios using cumulative reward metrics. The best-performing policies in each scenario are highlighted with a gray background.}
  \label{tab:performance_comparison}
\end{table}

\begin{figure}[htb]
    \centering
    \includegraphics[width=0.37\textwidth]{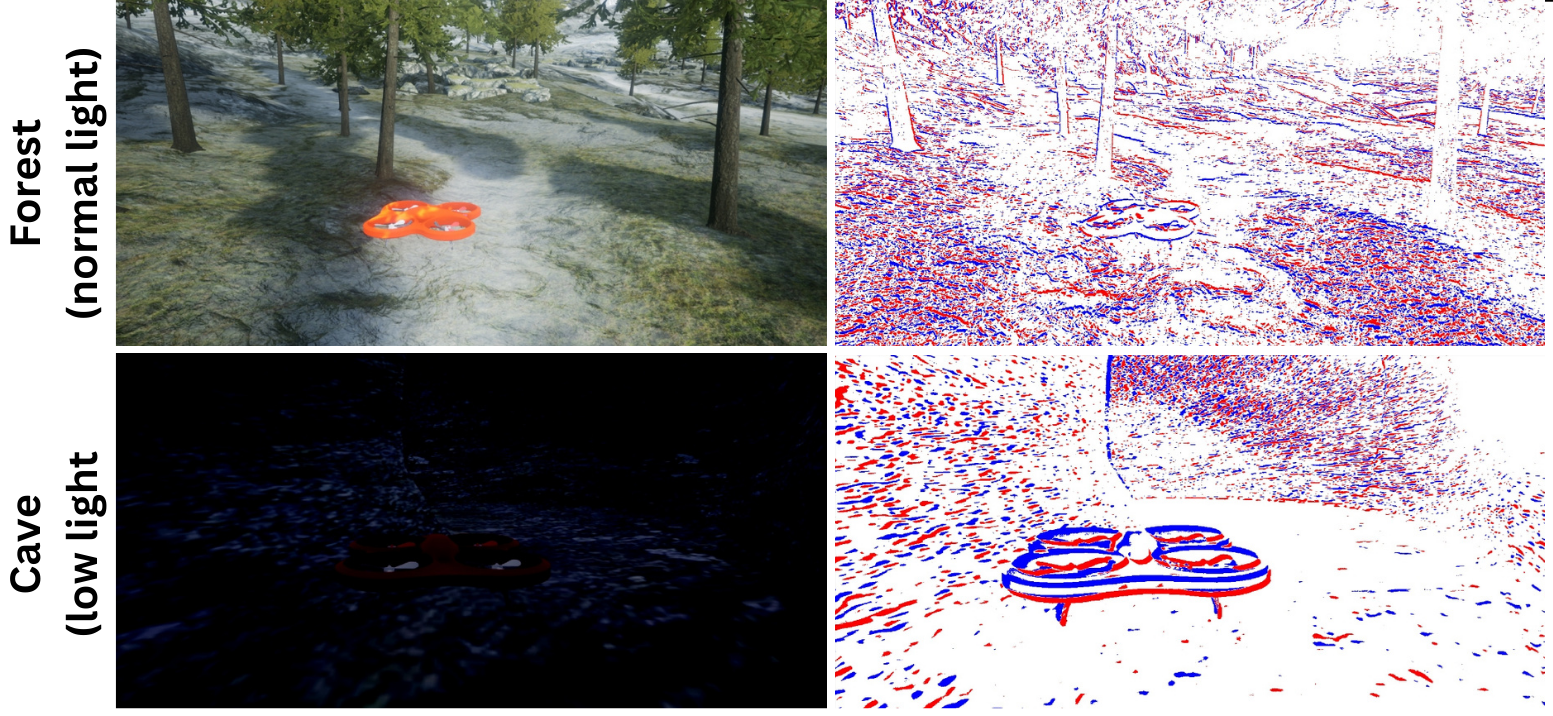}
    \caption{Illustration of tracking scenarios under normal and low light conditions, comparing RGB  and event-based inputs.}
    \label{fig:light}
\end{figure}

As depicted in the table, under low-light conditions, event cameras achieve the highest performance due to their high dynamic range, compared to the poor performance of underexposed RGB images in such scenarios.
In contrast,  light reflections on surfaces can cause significant pixel-level brightness changes in highly textured environments, leading to excessive noise from event cameras. In these cases, RGB cameras perform better. 
Two visual examples illustrating such situations are shown in Fig. \ref{fig:light}.

Additionally, in fast-target scenarios, RGB cameras with low frame rate can cause the target to move out of the frame quickly, leading to poor tracking performance. In contrast, event cameras capture high-frequency changes, enabling the drone to react to rapid movements with microsecond-level response times, resulting in superior performance for fast-moving targets as shown in the table. 
In summary, event cameras consistently  outperform RGB cameras, highlighting their advantages, except in high-textured environments where their performance is limited.

\section{Conclusion} 
\label{sec:conc}
In this paper, we addressed the problem of active drone tracking problem using  visual information from event cameras  and ASAC-based, deep reinforcement learning, highlighting the benefits of event cameras in terms of reactivity and robustness under varying conditions. We demonstrated the advantages of jointly optimizing the feature extractor and tracking process in an end-to-end fashion rather than training an object detector-based policy. The tracking policies were obtained via appropriate reward shaping within domain randomized environments in simplistic box-like environments to larger-scale, adverse conditions reminiscent of subterranean scenes.
A multi-modal solution that leverages the advantages of both  RGB  and event modalities could be a subject of further investigation, for example via hierarchical RL. Such extension, however, could require additional attention so as to not compromise real-time tracking performance.
\section{Acknowledgement}
This work has received a French government support granted to the Labex CominLabs excellence laboratory and managed by the National Research Agency in the ``Investing for the Future" program under reference ANR-10-LABX-07-01, project \textit{LEASARD} (Low-Energy deep neural networks for Autonomous 
Search-And-Rescue Drones).







\bibliographystyle{IEEEtran}
\bibliography{egbib}

\end{document}